\documentclass{iac}
\usepackage{textcomp}
\usepackage{subcaption}
\usepackage{hyperref}
\begin{document}

% \IACconference{76}
% \IAClocation{Sydney, Australia}
% \IACdates{29 Sept - 3 Oct}
% \IACyear{2025}
% \IACpapernumber{ D3, IPB, 17, x100422}
% \IACcopyright{2025}{the International Astronautical Federation (IAF)}

\title{RL-AVIST: Reinforcement Learning for Autonomous \\Visual Inspection of Space Targets}

% One affiliation, two authors, two emails
% Define affiliations once (with number IDs)
\IACaffiliation{Space Robotics Research Group, SnT, University of Luxermbourg}{matteo.elhariry@uni.lu}
\IACaffiliation{Earth Species Project}{}

% Declare authors and map them to affiliations using superscript indices
\IACauthor{Matteo El Hariry}{1}
\IACauthor{Andrej Orsula}{1}
\IACauthor{Matthieu Geist}{2}
\IACauthor{Miguel Olivares Mendez}{1}

\abstract{The growing need for autonomous on-orbit services such as inspection, maintenance, and situational awareness calls for intelligent spacecraft capable of complex maneuvers around large orbital targets. Traditional control systems often fall short in adaptability, especially under model uncertainties, multi-spacecraft configurations, or dynamically evolving mission contexts. This paper introduces RL-AVIST, a Reinforcement Learning framework for Autonomous Visual Inspection of Space Targets. Leveraging the Space Robotics Bench (SRB), we simulate high-fidelity 6-DOF spacecraft dynamics and train agents using DreamerV3, a state-of-the-art model-based RL algorithm, with PPO and TD3 as model-free baselines. Our investigation focuses on 3D proximity maneuvering tasks around targets such as the Lunar Gateway and other space assets. We evaluate task performance under two complementary regimes: generalized agents trained on randomized velocity vectors, and specialized agents trained to follow fixed trajectories emulating known inspection orbits. Furthermore, we assess the robustness and generalization of policies across multiple spacecraft morphologies and mission domains. Results demonstrate that model-based RL offers promising capabilities in trajectory fidelity, and sample efficiency, paving the way for scalable, retrainable control solutions for future space operations.}
\maketitle

\section{Introduction}

The next generation of space exploration missions will require increased levels of autonomy for spacecraft operating in complex orbital environments~\cite{Nesnas2021}. Key mission capabilities, such as structural inspection, proximity operations, and situational awareness, are often constrained by communication delays, model uncertainties, and dynamically changing conditions in the space environment~\cite{Banerjee2023}. To meet these challenges, intelligent control architectures capable of real-time adaptation and generalization are a promising solution.

Traditional controllers, based on analytical models and hand-tuned parameters, struggle to scale with such demands. Recent advances in reinforcement learning (RL) offer a promising alternative: instead of relying on explicit modeling, an agent learns to infer effective control policies through interaction with its environment. This is especially beneficial in proximity operations, where unmodeled disturbances or fuel constraints require a high degree of flexibility. Model-free RL has already demonstrated viability in tasks such as waypoint tracking on rough terrain~\cite{Orsula2025Sim2Dust}, but remains sample-inefficient and brittle when generalization across multiple vehicle configurations is required.

\begin{figure}[t]
% \captionsetup{font=small}  % sets caption font to small
    \centering      
    \includegraphics[width=.99\linewidth ]{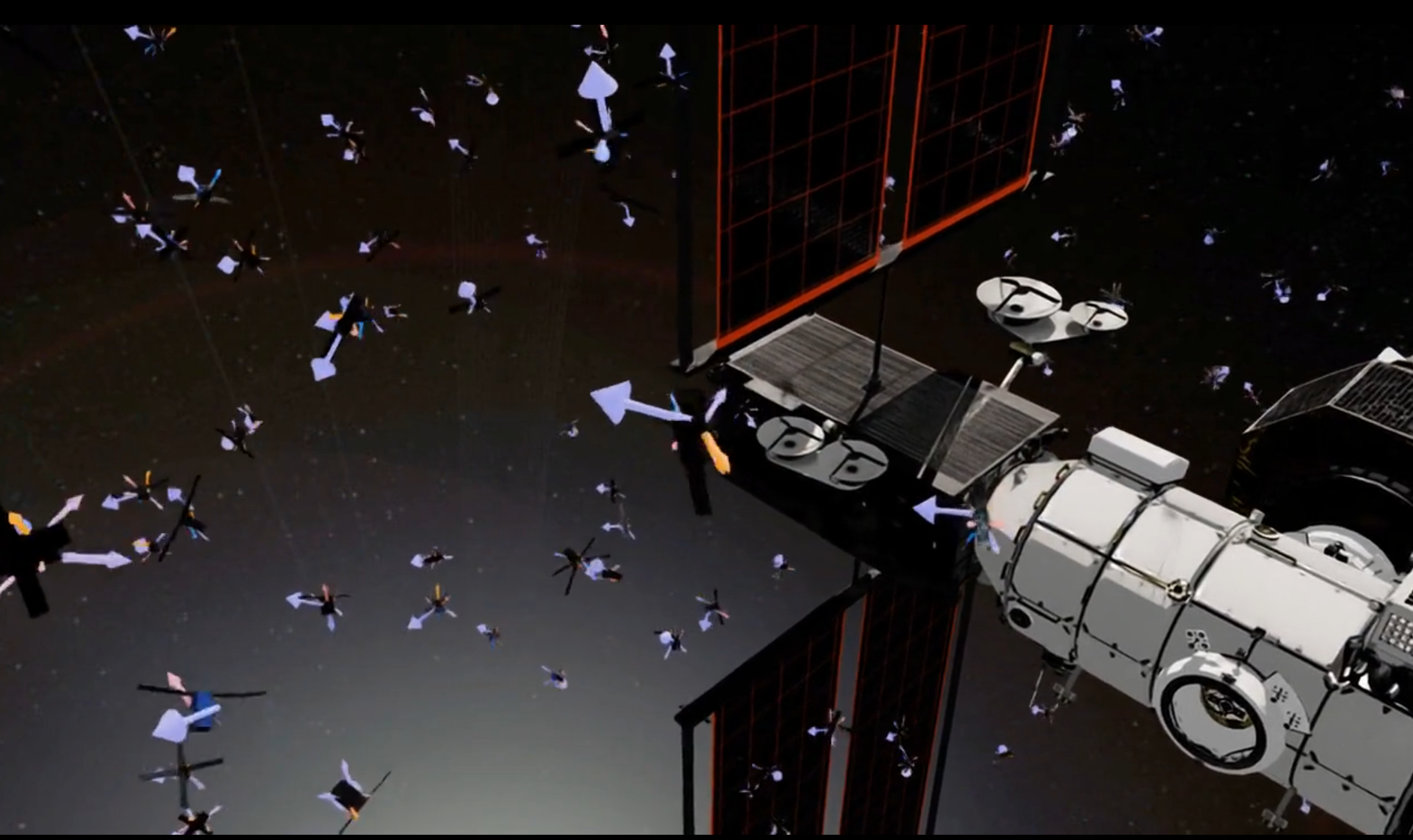}
    \small
    \caption{Training of multiple CubeSat morphologies to follow randomized velocity vectors in the vicinity of the Lunar Gateway. The target structure is shown for visual context only, while the agents learn generalist control policies through diverse randomized conditions.}
    \label{fig:banner}
\end{figure}

In this work, we present RL-AVIST, Reinforcement Learning for Autonomous Visual Inspection of Space Targets. The framework is built on the Space Robotics Bench (SRB)~\cite{Orsula2024SRB}, enabling high-fidelity simulation of spacecraft dynamics with support for multiple morphologies and inspection targets. Fig.~\ref{fig:banner}, displays different cubesats spawned closeby the Lunar Gateway target during a training episode. Unlike prior work focused on ground rovers or 2D proximity operations, RL-AVIST operates in 3D and models continuous-thrust dynamics suited for spacecraft maneuvering.

Our approach leverages Dreamer, a model-based RL algorithm~\cite{Hafner2020Dreamer}, to learn compact latent-space models of system dynamics, enabling efficient trajectory planning and control. We compare DreamerV3 with strong model-free baselines such as Proximal Policy Optimization (PPO)~\cite{Schulman2017PPO} and Twin Delayed DDPG (TD3)~\cite{Fujimoto2018TD3} across multiple inspection tasks. Agents are trained under two main regimes: one that exposes them to randomized velocity commands to test general adaptability, and another that trains them on specific orbital patterns for structured inspection missions. To evaluate generalization, we further test policies on unseen spacecraft models and orbital assets such as the Lunar Gateway~\cite{Lehnhardt2024}.

Through a series of ablation studies and evaluations, we demonstrate that model-based RL significantly improves sample efficiency and while maintaining trajectory accuracy. These results represent a step forward in scalable autonomy for on-orbit operations and contribute to the vision of future space infrastructure managed by intelligent agents.

\section{Related Work}

Autonomous spacecraft inspection is an increasingly active research area, driven by the need for in-orbit servicing, structural monitoring, and future on-orbit assembly. Early inspection strategies often relied on pre-programmed motion plans and highly specialized control architectures~\cite{Pedrotty2020}, with systems such as NASA's Seeker demonstrating short-range free-flyer navigation capabilities. These approaches, while effective for specific mission profiles, remain rigid in the face of evolving operational contexts or varying satellite geometries.

More recent efforts explore the deployment of compact, autonomous platforms for inspection missions. ~\cite{Corpino2020} investigate the use of multi-purpose CubeSats to inspect the Lunar Gateway, analyzing system trade-offs across power, sensing, and propulsion. Similarly,~\cite{Nakka2022} propose an information-based guidance and control architecture enabling multi-spacecraft collaborative inspection of large assets. These approaches begin to address the generalization and scalability needed for long-duration autonomous operations in orbit.

Reinforcement learning (RL) has emerged as a promising technique to equip spacecraft with adaptable control strategies that do not require precise analytical models. Model-free RL methods such as PPO have shown success in robotic navigation and waypoint tracking, including demonstration of terrain-adaptive policies for rovers navigating deformable surfaces~\cite{Orsula2025Sim2Dust}. This work was built on the Space Robotics Bench (SRB)~\cite{Orsula2024SRB}, a simulation platform for benchmarking learning-based control across diverse planetary and orbital scenarios.

In the orbital domain, the "SmallSat Steward" system~\cite{Majumdar2025} was introduced using a reinforcement learning architecture for reactive close-proximity operations. By combining direct RL with online model learning, their method achieves adaptability to changing thruster dynamics and system uncertainties, though their work is limited to two-dimensional scenarios. Complementary efforts have demonstrated RL in increasingly realistic docking and inspection contexts: ~\cite{hovell2021deep} combined deep RL with conventional control for transferable proximity guidance; ~\cite{oestreich2021autonomous} applied PPO to enable 6-DOF docking with rotating targets; and ~\cite{aborizk2024multiphase} proposed a hierarchical model-based RL approach for multiphase docking under complex constraints. In parallel, autonomy frameworks such as the TumbleDock flight experiment~\cite{albee2022autonomous} and microgravity free-flyer inspection architectures~\cite{albee2025architecting} showcase the importance of safety, replanning, and real-world validation.  In a 3 degrees-of-freedom environment, DRIFT~\cite{el2024drift} provides a sim-to-real DRL demonstration for floating platforms control, comparing DRL and optimal control under different levels of uncertainty. ~\cite{Iversflaten2025} further highlight adaptability with a robust replanning strategy for multi-agent inspection.

To address sample inefficiency and enhance planning capabilities, model-based RL methods such as DreamerV3~\cite{Hafner2020Dreamer} have been proposed. These methods learn compact latent models of the environment to support imagination-based policy learning, enabling sample-efficient planning in complex dynamics. While DreamerV3 has been widely studied in terrestrial tasks, its use in spacecraft control, particularly for visual inspection in 6-DOF orbital dynamics, remains largely unexplored.

This work builds upon these foundations by extending model-based reinforcement learning to the domain of orbital inspection. We propose a learning framework that supports continuous-thrust dynamics, varying spacecraft morphologies, and multiple mission targets such as the Lunar Gateway~\cite{Lehnhardt2024}. By leveraging the SRB platform, we aim to systematically evaluate policy generalization, robustness, and performance across diverse orbital scenarios, contributing toward scalable and retrainable control strategies for future autonomous space infrastructure.

\section{RL-AVIST Framework}

The RL-AVIST framework is designed to enable learning-based autonomous visual inspection of space assets in simulation, with a focus on adaptability, generalization, and precision under realistic spacecraft dynamics. It builds upon the Space Robotics Bench (SRB)~\cite{Orsula2024SRB}, extending it with spacecraft-specific modules for visual inspection scenarios and dynamic target broadcasting.

\subsection*{Problem Formulation}

Each episode in the RL-AVIST environment simulates a 6-DOF free-flying spacecraft initialized at a randomized pose relative to a target object, such as a satellite or station module. The control task is modeled as a Partially Observable Markov Decision Process (POMDP), defined by the tuple $\mathcal{M} = \langle \mathcal{S}, \mathcal{A}, \mathcal{O}, T, R, \gamma \rangle$, where $\mathcal{S}$ is the true state space, $\mathcal{A} \subset \mathbb{R}^8$ is the continuous action space representing normalized thrust levels of 8 body-fixed thrusters, $\mathcal{O}$ is the observation space, $T$ is the transition function induced by spacecraft dynamics, $R$ is the reward function, and $\gamma$ is the discount factor.

At each timestep $t$, the agent receives an observation vector $o_t \in \mathcal{O}$ composed of:
\[
o_t = \left[\mathbf{a}_{t-1},\ \mathbf{v}_{\text{lin}},\ \boldsymbol{\omega}_{\text{ang}},\ \Delta \mathbf{p},\ \Delta \mathbf{R}_{6} \right]
\]
where $\mathbf{a}_{t-1} \in \mathbb{R}^8$ is the previous thruster command, $\mathbf{v}_{\text{lin}}, \boldsymbol{\omega}_{\text{ang}} \in \mathbb{R}^3$ are the current linear and angular velocities of the spacecraft, and $\Delta \mathbf{p} \in \mathbb{R}^3$, $\Delta \mathbf{R}_{6} \in \mathbb{R}^6$ denote the relative position and orientation (6D representation) to the target in the body frame.

The agent outputs a continuous action vector $\mathbf{a}_t \in \mathbb{R}^8$, where each element corresponds to the activation level of a directional thruster mounted at a fixed offset from the spacecraft’s center of mass. The resulting forces and torques are computed via:
\[
\mathbf{F} = \sum_{i=1}^{8} a_i P_i \mathbf{d}_i, \quad \boldsymbol{\tau} = \sum_{i=1}^{8} \mathbf{r}_i \times (a_i P_i \mathbf{d}_i)
\]
where $P_i$ is the maximum thrust power, $\mathbf{d}_i$ is the unit direction vector, and $\mathbf{r}_i$ the body-frame offset of the $i$-th thruster.

The reward signal $r_t$ is designed to guide the agent toward smooth, efficient, and precise relative navigation. It includes the following components:

- \textbf{Control penalties:} Action magnitude and rate are penalized to encourage energy-efficient and smooth actuation, with terms proportional to $\|\mathbf{a}_t\|^2$ and $\|\mathbf{a}_t - \mathbf{a}_{t-1}\|^2$ respectively.

- \textbf{Position tracking:} A quadratic penalty term proportional to $\|\Delta \mathbf{p}\|^2$ discourages large distances, while a shaped reward $\left(1 - \tanh(\|\Delta \mathbf{p}\| / \sigma_p)\right)$ promotes convergence to the target.

- \textbf{Orientation alignment:} Once within close proximity, the agent is rewarded for minimizing the Frobenius norm of the rotation error matrix $\|\Delta \mathbf{R} - \mathbf{I}\|_F$, shaped by $\left(1 - \tanh(\|\cdot\| / \sigma_o)\right)$.

- \textbf{Stability at target:} To promote controlled and precise arrival, an additional bonus is applied for minimizing action rate at target proximity and alignment.

The total reward is thus expressed as:
\[
r_t = \lambda_1 r_{\text{track}} + \lambda_2 r_{\text{align}} + \lambda_3 r_{\text{stable}} + \lambda_4 p_{\text{mag}} + \lambda_5 p_{\text{rate}}
\]
where the $\lambda_i$ are scalar weights used to balance the influence of each term.

This design encourages the agent to reach the target position and orientation precisely, with minimal fuel expenditure and high stability, critical properties for on-orbit inspection and servicing tasks.

\subsection*{Task definition}

The target structure, such as the Lunar Gateway, ISS, or a Venus-orbiting satellite~\cite{Svedhem2009VenusExpress}, is assumed to be quasi-static in the relative reference frame, i.e., it exhibits negligible relative velocity with respect to the inspecting spacecraft. For inspection planning, the structure may be modeled as either fixed in space or following a known reference trajectory, which is broadcast at runtime via a dedicated module (ROS TF). Inspection paths are defined either by precomputed geometric patterns (e.g., circular, capsule, leminscale) or by dynamically streamed velocity vectors representing adaptive, task-driven behavior.

\subsection{Control policies and training}
We support both model-based and model-free reinforcement learning (RL) algorithms, with Dreamer~\cite{Hafner2020Dreamer} serving as the primary learning architecture. DreamerV3 learns a compact latent world model of spacecraft dynamics, enabling planning through imagination rollouts within the latent space. This allows for efficient credit assignment over long horizons and accelerates policy convergence. For comparison, we implement alternative model-free baselines using PPO and TD3, covering a spectrum of policy classes.

Training is conducted via the SRB control pipeline, using Hydra configuration files and a PyTorch-based Gym backend. All agents are trained in simulation with episodic resets and randomized domain parameters, such as initial position, target velocity, and spacecraft morphology, applied independently across parallel environments. Policies are evaluated on both dynamically streamed goal velocities and fixed geometric inspection trajectories. Key performance metrics include position and orientation error, control jerk, and smoothness.

To evaluate generalization, agents are trained under two complementary regimes. The first uses randomized goal velocities to promote flexible trajectory tracking across unseen scenarios. The second follows known reference trajectories, reflecting structured inspection patterns around mission-critical targets. Additionally, we compare agents trained on a single spacecraft model against those trained on multiple morphologies, assessing transferability across vehicle configurations. Optional disturbance injections, such as force or torque perturbations, can be activated to evaluate policy robustness.

This modular framework supports extensive experimentation across learning architectures, trajectory generation strategies, spacecraft designs, and target environments. The full codebase is integrated into SRB, with reusable modules for trajectory generation, control, and performance evaluation.

\begin{figure}[t]
    \centering
    \includegraphics[width=0.95\linewidth]{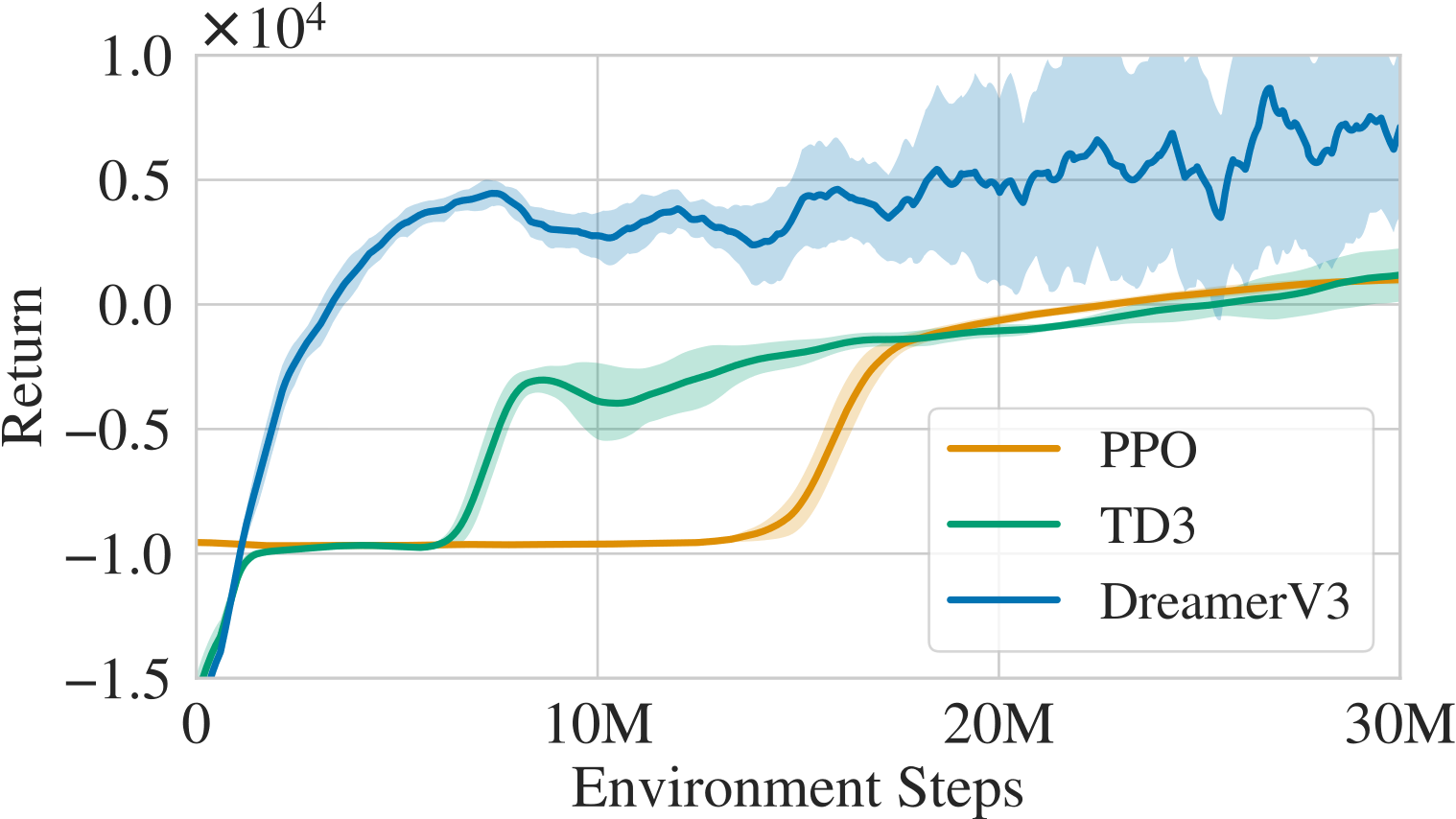}
    \caption{Training performance (mean ± std across 3 seeds) for Dreamer, PPO, and TD3 on the generalist randomized goal-velocity task.}
    \label{fig:learning-curves}
\end{figure}

\section{Experiments \& Results}

\begin{figure*}[htbp]
    \centering
    % First row
    \subfloat[Capsule]{%
        \includegraphics[width=0.3\linewidth]{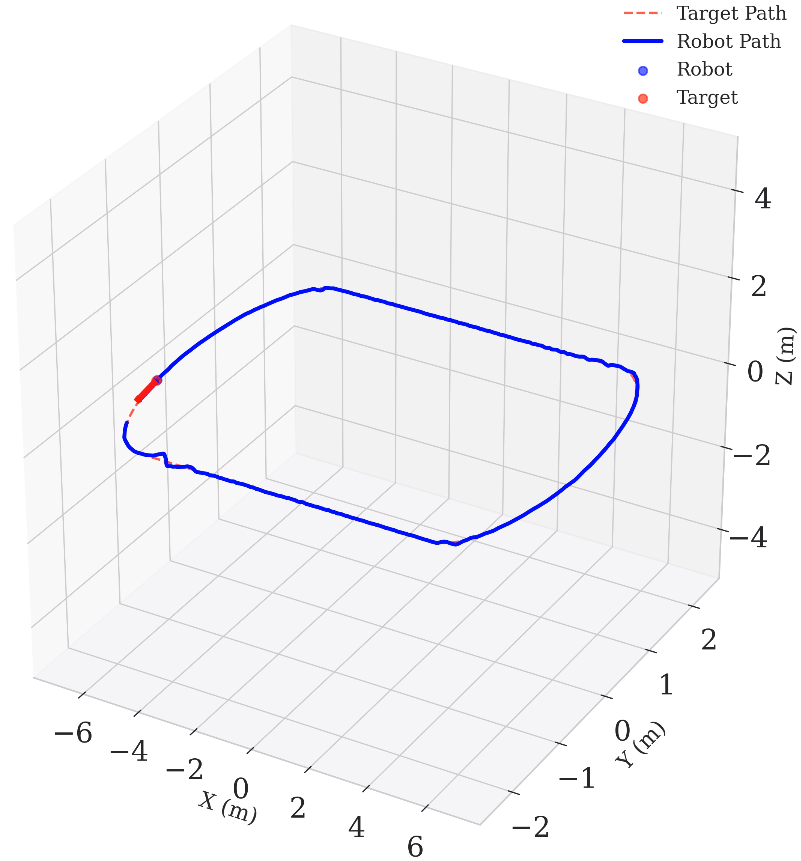}
    } \hspace{4pt}
    \subfloat[Circle]{%
        \includegraphics[width=0.3\linewidth]{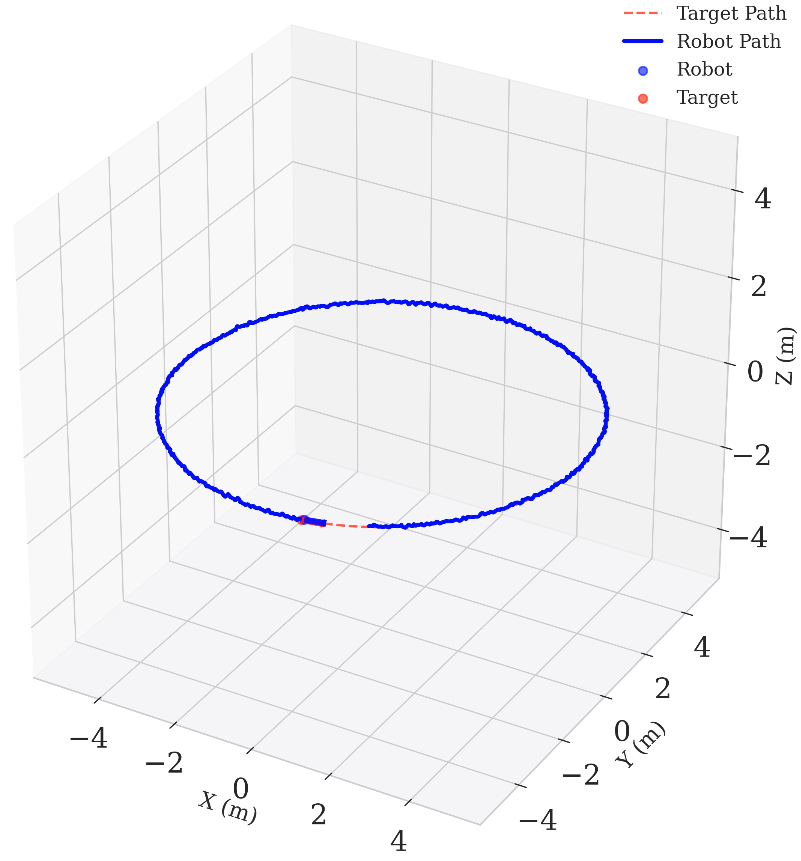}
    } \hspace{4pt}
    \subfloat[Rectangle]{%
        \includegraphics[width=0.3\linewidth]{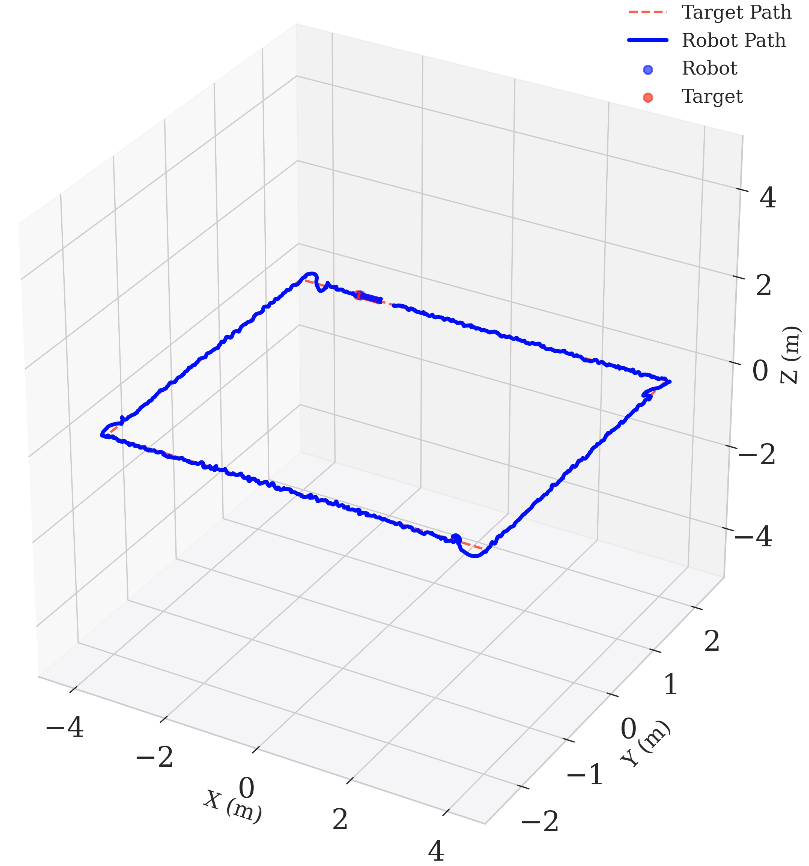}
    }\\
    \subfloat[Lemniscate]{%
        \includegraphics[width=0.3\linewidth]{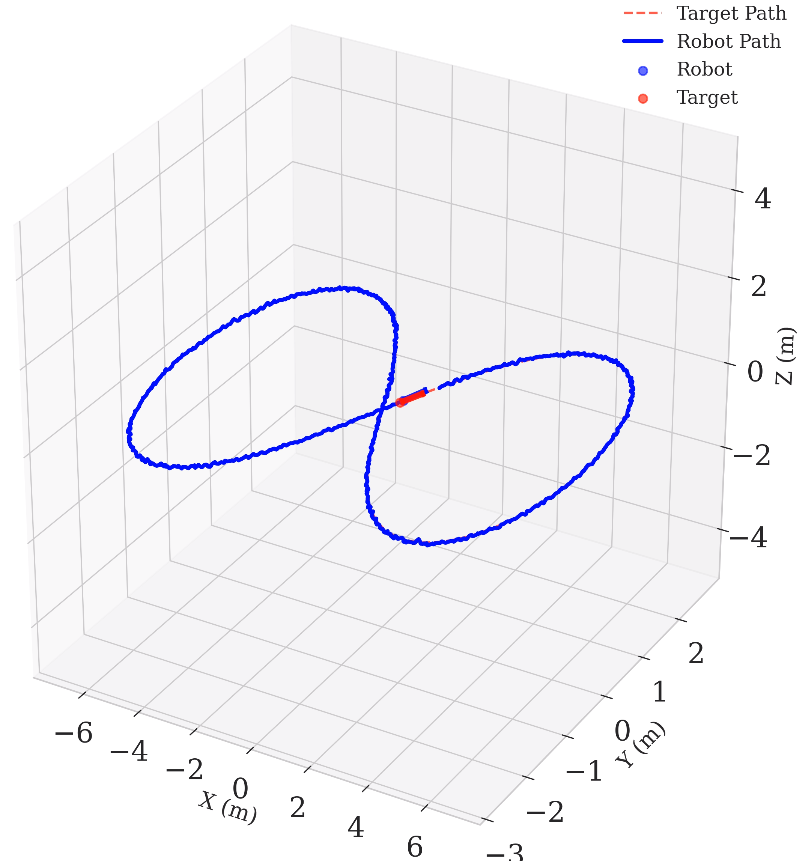}
    } \hspace{4pt}
    \subfloat[Lissajous]{%
        \includegraphics[width=0.3\linewidth]{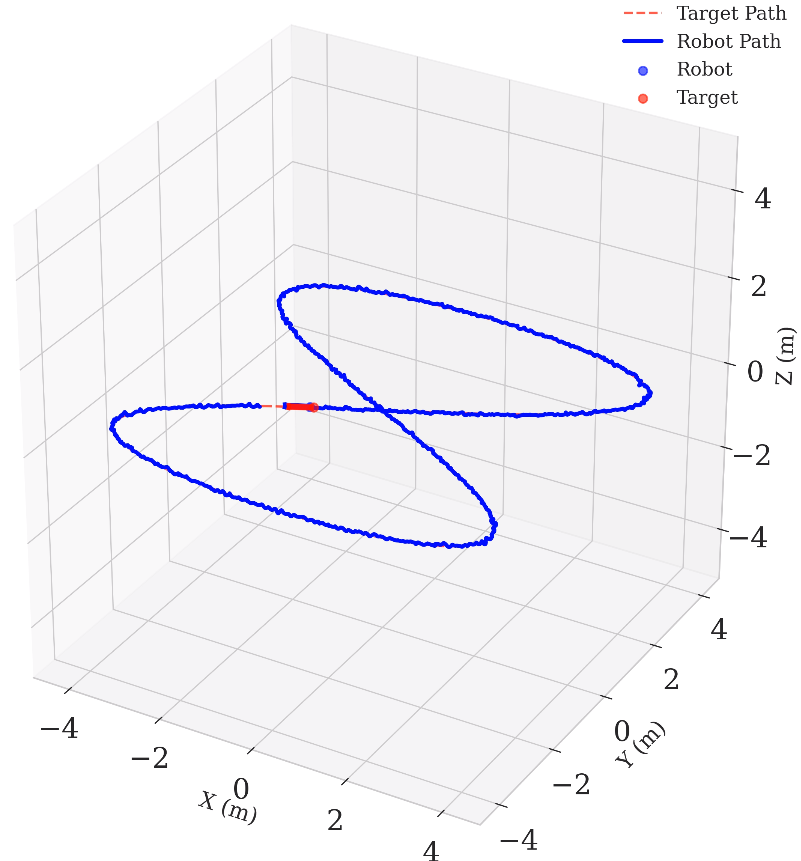}
    } \hspace{4pt}
    \subfloat[Spiral]{%
        \includegraphics[width=0.3\linewidth]{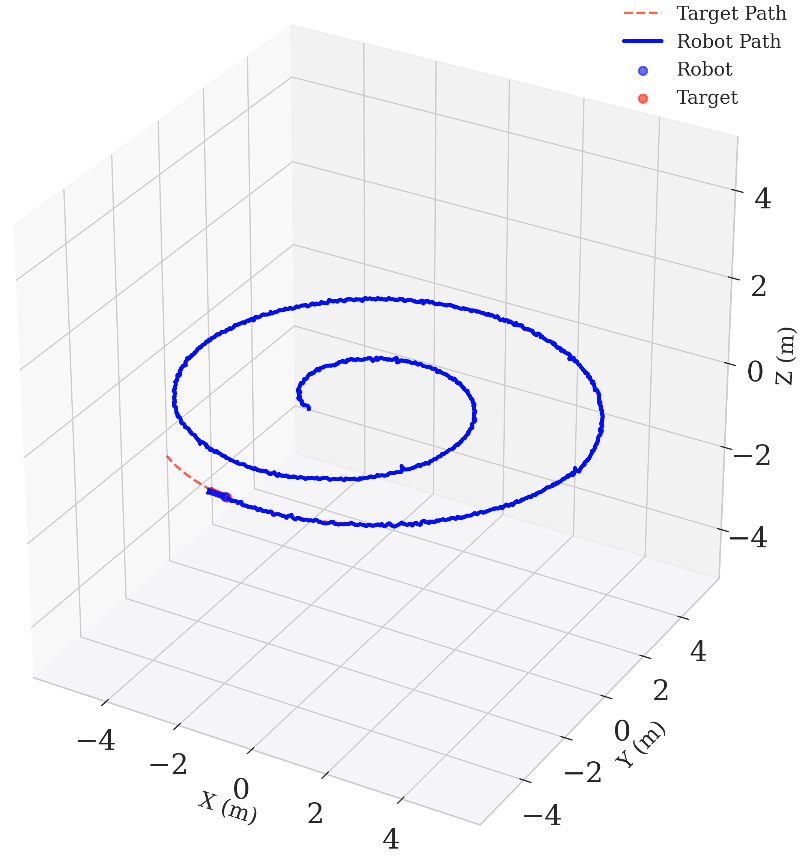}
    }
    % \subfloat[]{%
    %     \includegraphics[width=0.24\linewidth]{figures/placeholder.png}
    % }\hfill
    % \subfloat[]{%
    %     \includegraphics[width=0.24\linewidth]{figures/placeholder.png}
    % }
    \caption{3D tracking performance of the DreamerV3 policy across multiple inspection trajectories: capsule, circle, rectangle, lemniscate, Lissajous, and spiral.}
    \label{fig:traj_shapes}
\end{figure*}

We evaluate the RL-AVIST pipeline through a series of experiments designed to analyze generalization, specialization, and real-world applicability in orbital visual inspection tasks.

\subsection{Comparison of RL Algorithms on Generalist Training}

In the first set of experiments, we train generalist policies using three different RL algorithms: Dreamer~\cite{Hafner2020Dreamer}, PPO~\cite{Schulman2017PPO}, and TD3~\cite{Fujimoto2018TD3}. Each agent is trained to follow randomized velocity vectors across randomized spacecraft morphologies. Results are averaged over three random seeds per algorithm.

As shown in Fig.~\ref{fig:learning-curves}, DreamerV3 achieves the highest returns and fastest convergence, clearly outperforming PPO and TD3. TD3 shows moderate success but slower learning, while PPO lags significantly behind. The superior performance of DreamerV3 highlights the advantage of model-based RL for long-horizon planning and generalization across spacecraft variations.

\subsection{Testing Across Diverse Inspection Trajectories}

After confirming in the previous section that DreamerV3 consistently outperforms PPO and TD3, we select it as the reference model for further evaluation. To assess its versatility, we deploy the DreamerV3 policy on a diverse set of inspection trajectories, including capsule, circle, rectangle, lemniscate, Lissajous, and spiral paths. These trajectories represent both structured inspection maneuvers and more complex geometric patterns.

Figure~\ref{fig:traj_shapes} illustrates the resulting 3D tracking performance. Across all tested trajectories, DreamerV3 is able to maintain smooth control and accurate path following, demonstrating strong adaptability to varying inspection patterns relevant for orbital scenarios.

\subsection{Visual Inspection Around Realistic Assets}

\begin{figure*}[t]
\vspace{-15pt}
    \centering
    \subfloat{%
        \includegraphics[width=0.9\linewidth,height=0.4\linewidth]{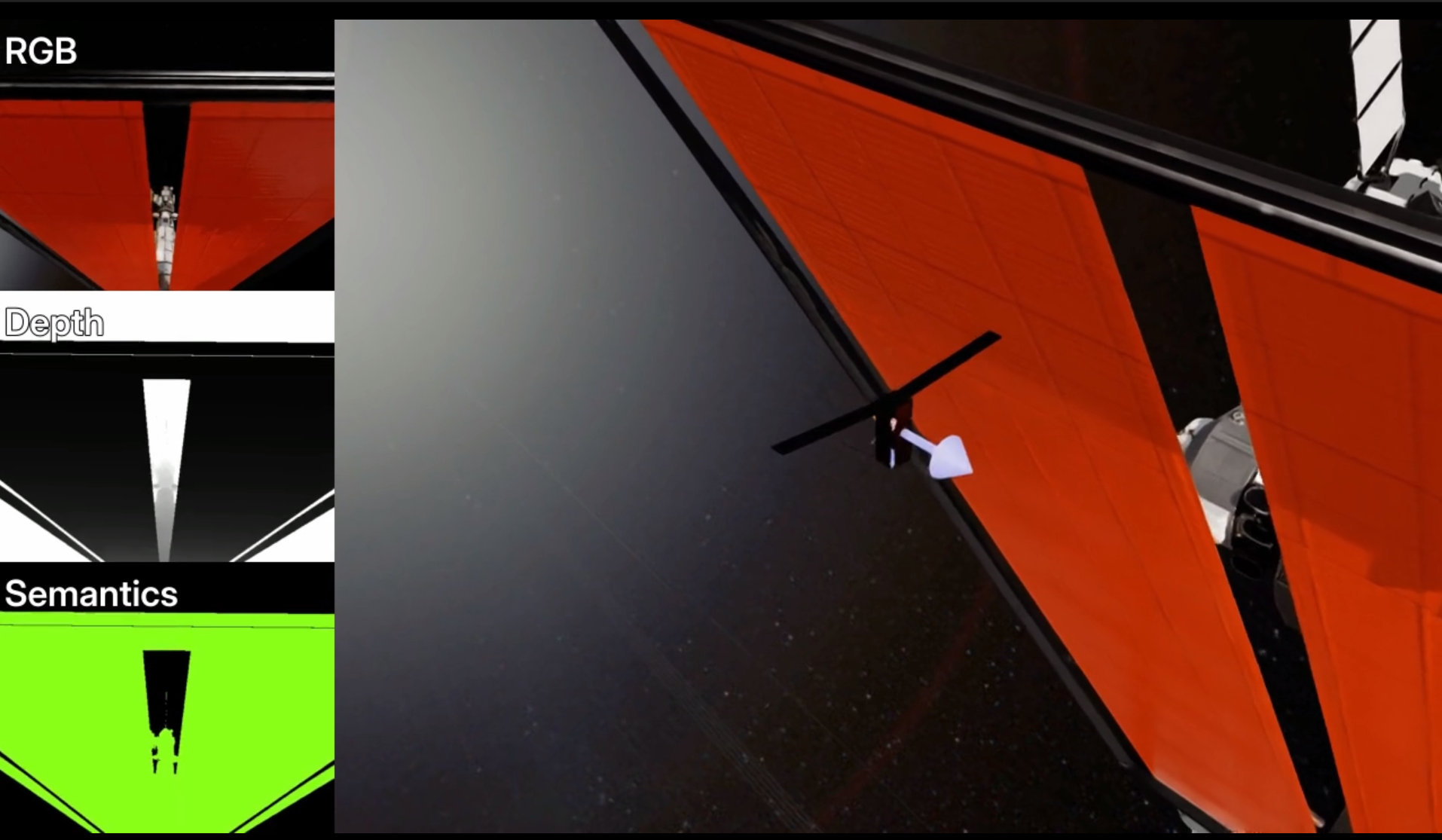}
    }\hfill
    \subfloat{%
         \includegraphics[width=0.9\linewidth,height=0.4\linewidth]{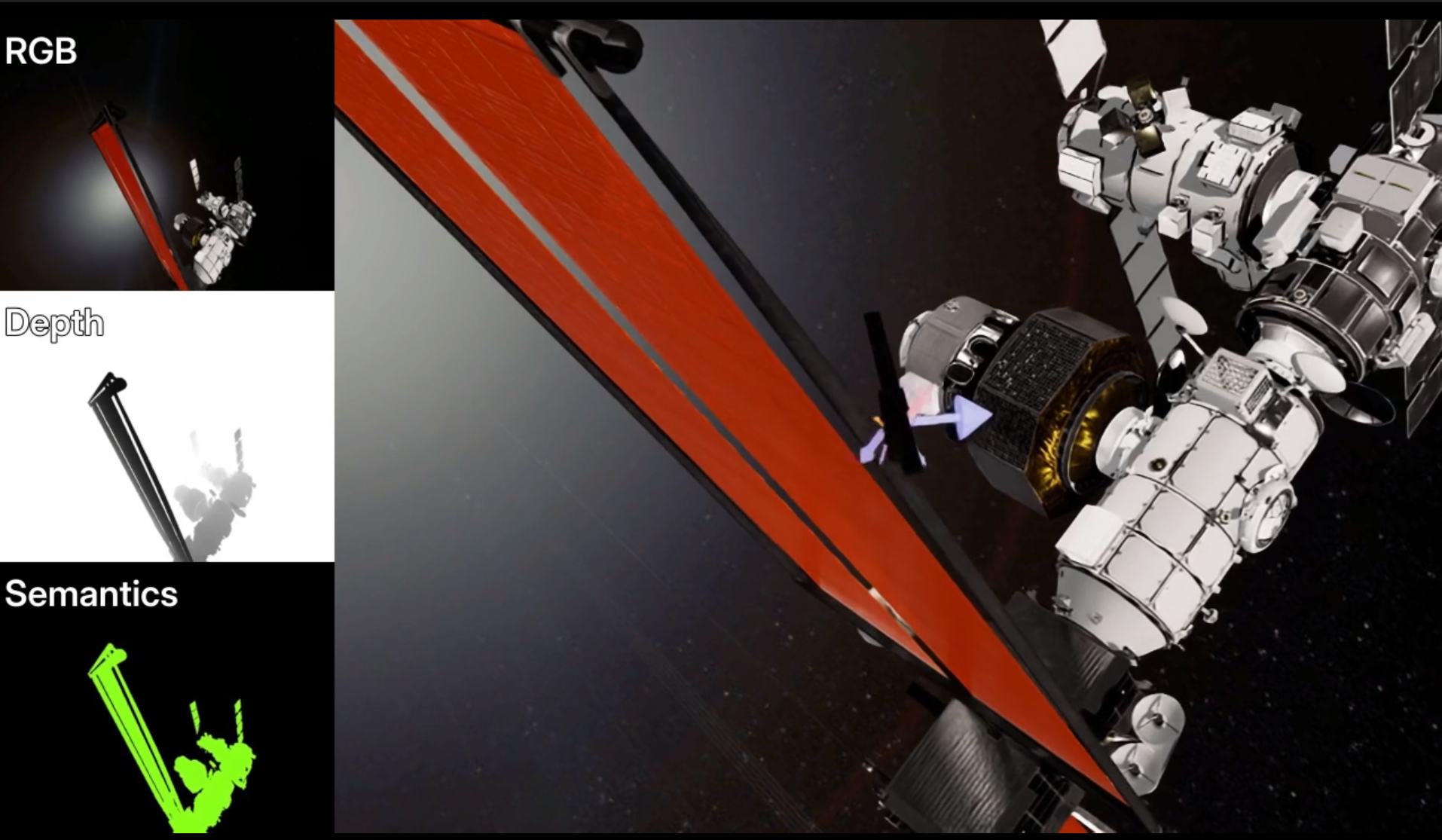}
    }\hfill
    \subfloat{%
         \includegraphics[width=0.9\linewidth,height=0.4\linewidth]{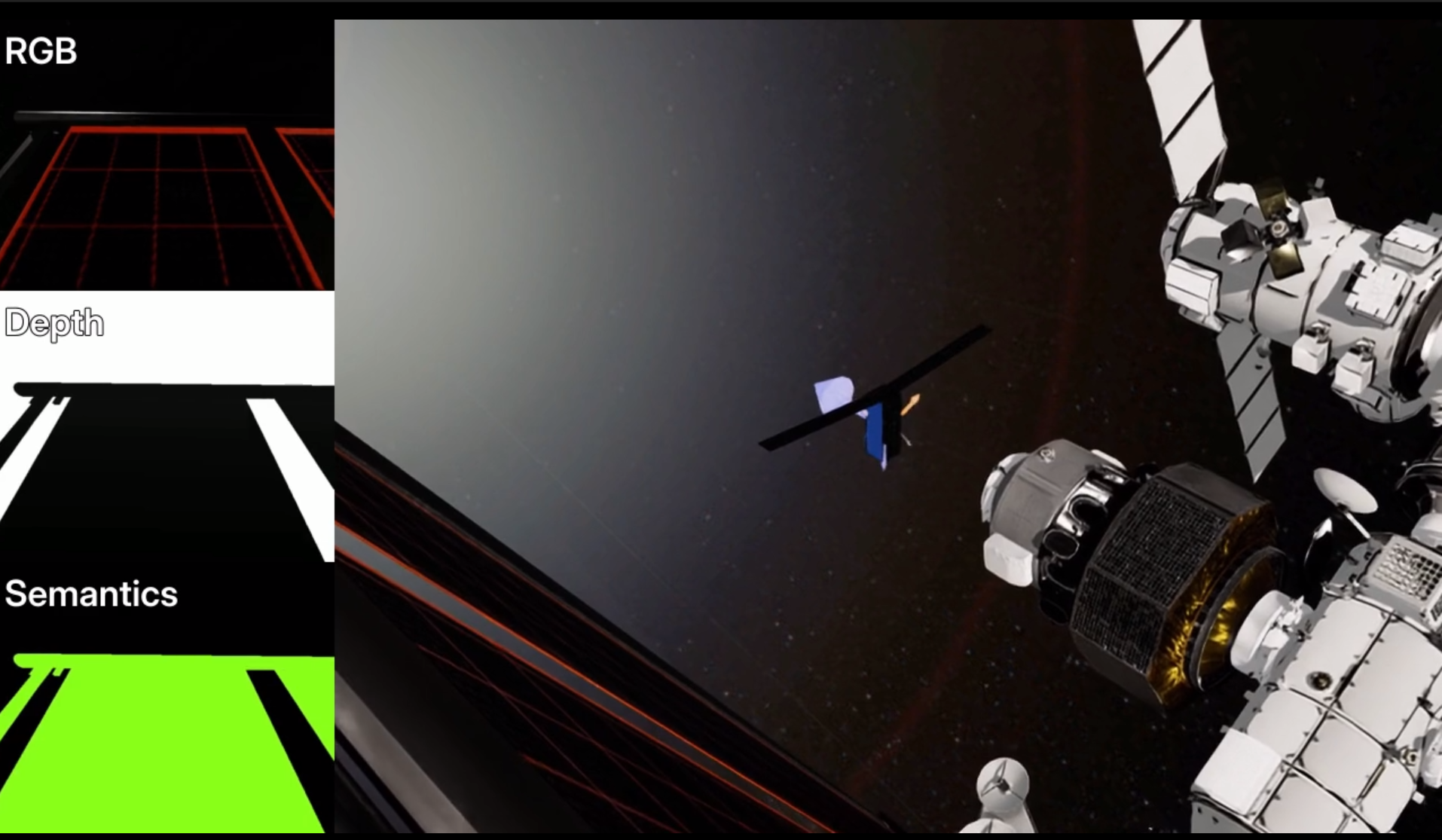}
    }
    \small
    \caption{Sequence of inspection frames showing the spacecraft trajectory around the Lunar Gateway. Each frame illustrates RGB, depth, and semantic views captured during different stages of the maneuver, highlighting accurate tracking and consistent perception throughout the inspection.}

    \label{fig:traj_sequence_gateway}
\end{figure*}

\begin{figure*}[t]
\vspace{-15pt}
    \centering
    \subfloat{%
        \includegraphics[width=0.9\linewidth,height=0.4\linewidth]{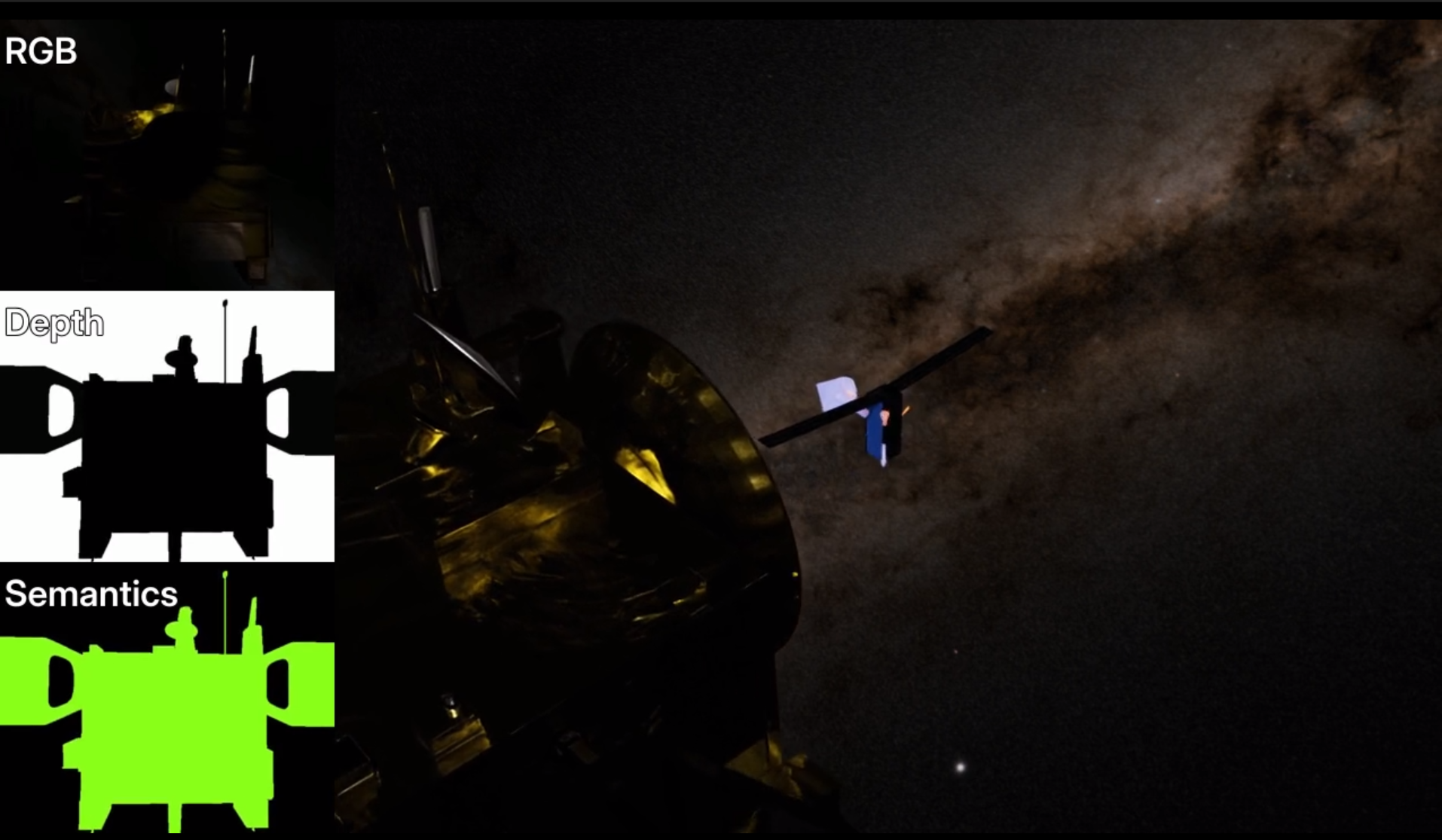}
    }\hfill
    \subfloat{%
         \includegraphics[width=0.9\linewidth,height=0.4\linewidth]{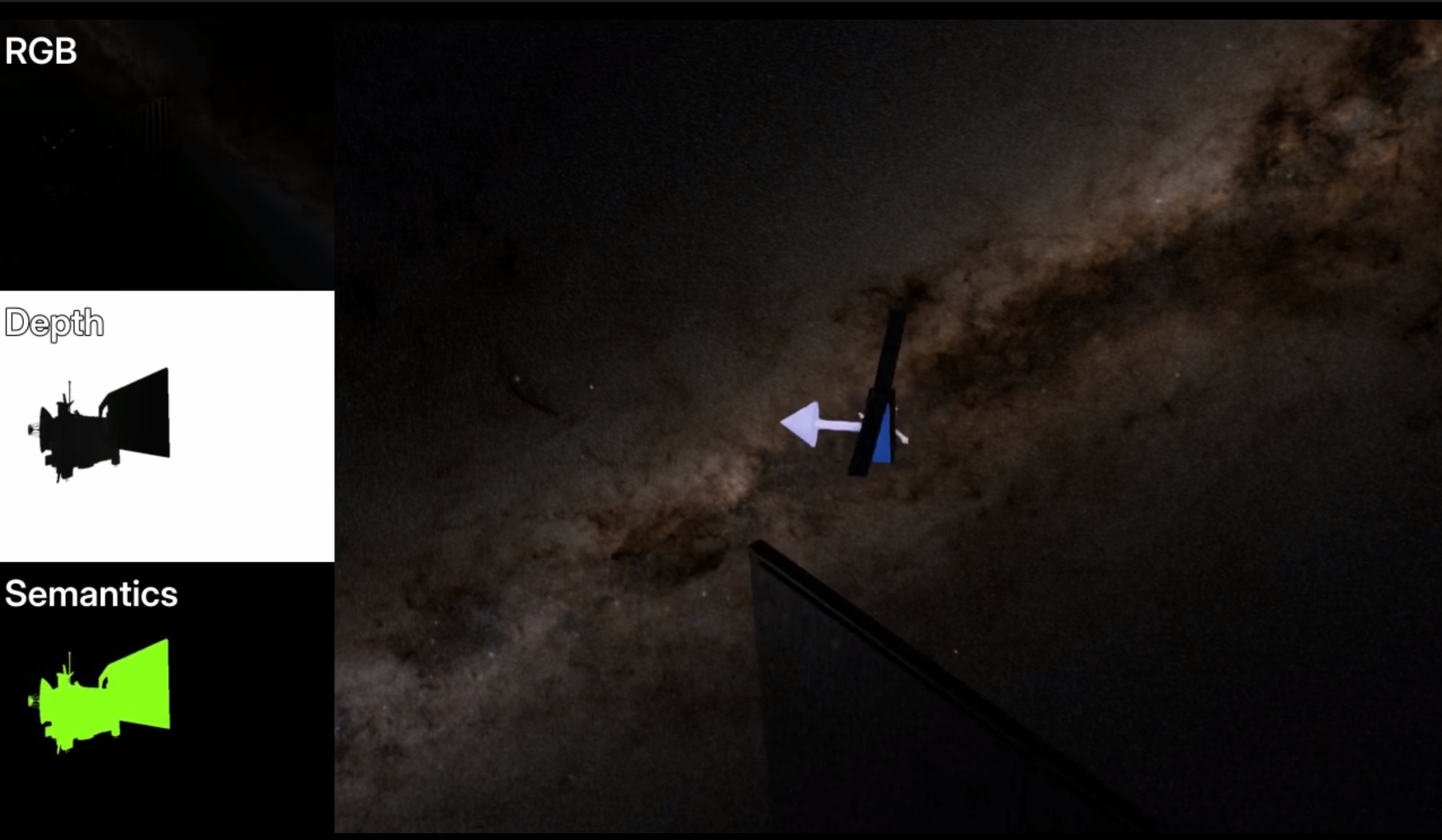}
    }\hfill
    \subfloat{%
         \includegraphics[width=0.9\linewidth,height=0.4\linewidth]{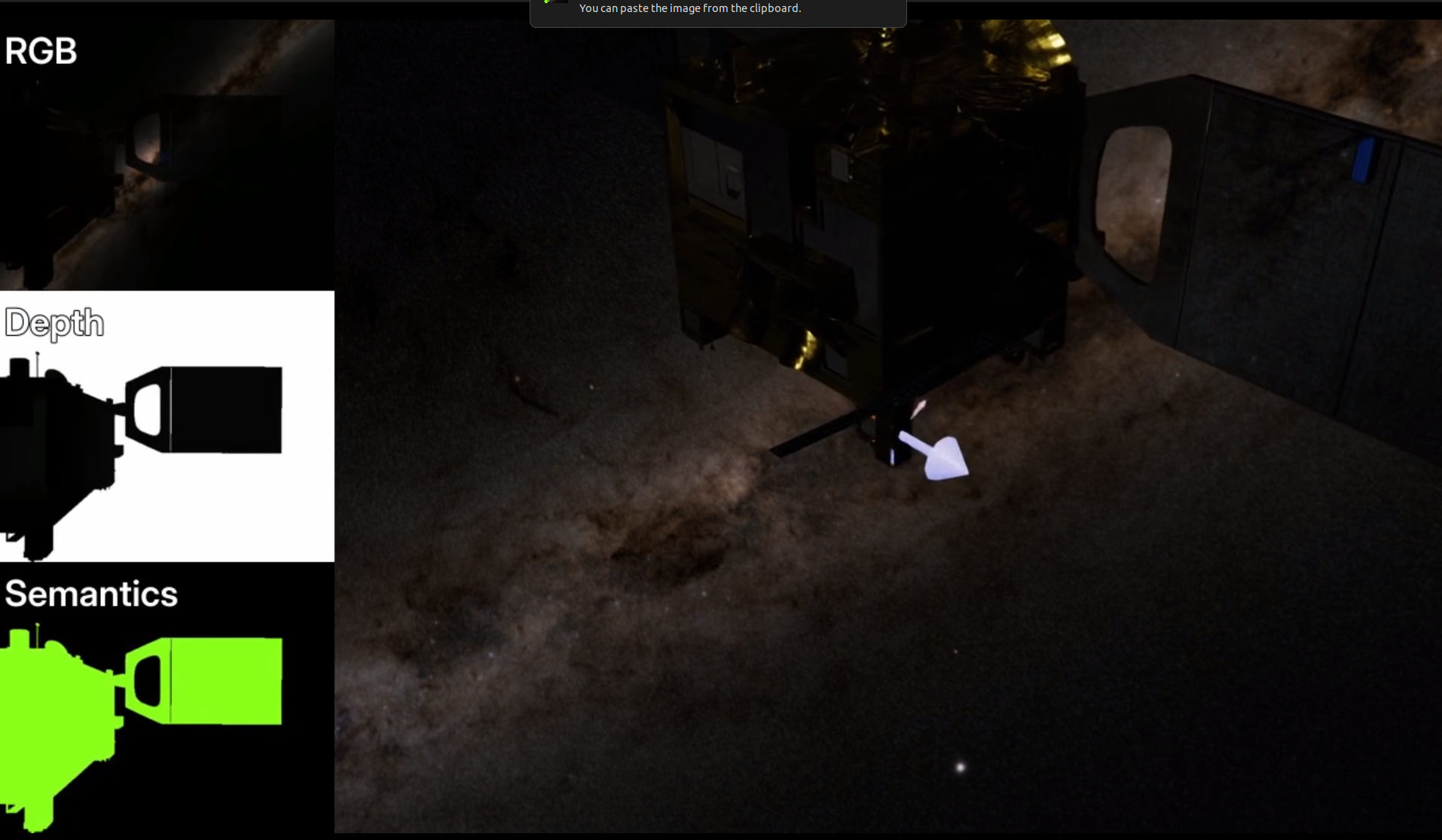}
    }
    \small
    \caption{Sequence of inspection frames showing the spacecraft trajectory around the Venus Express probe.}

    \label{fig:traj_sequence_venus}
\end{figure*}

\begin{figure*}[t]
\vspace{-15pt}
    \centering
    \subfloat{%
        \includegraphics[width=0.9\linewidth,height=0.4\linewidth]{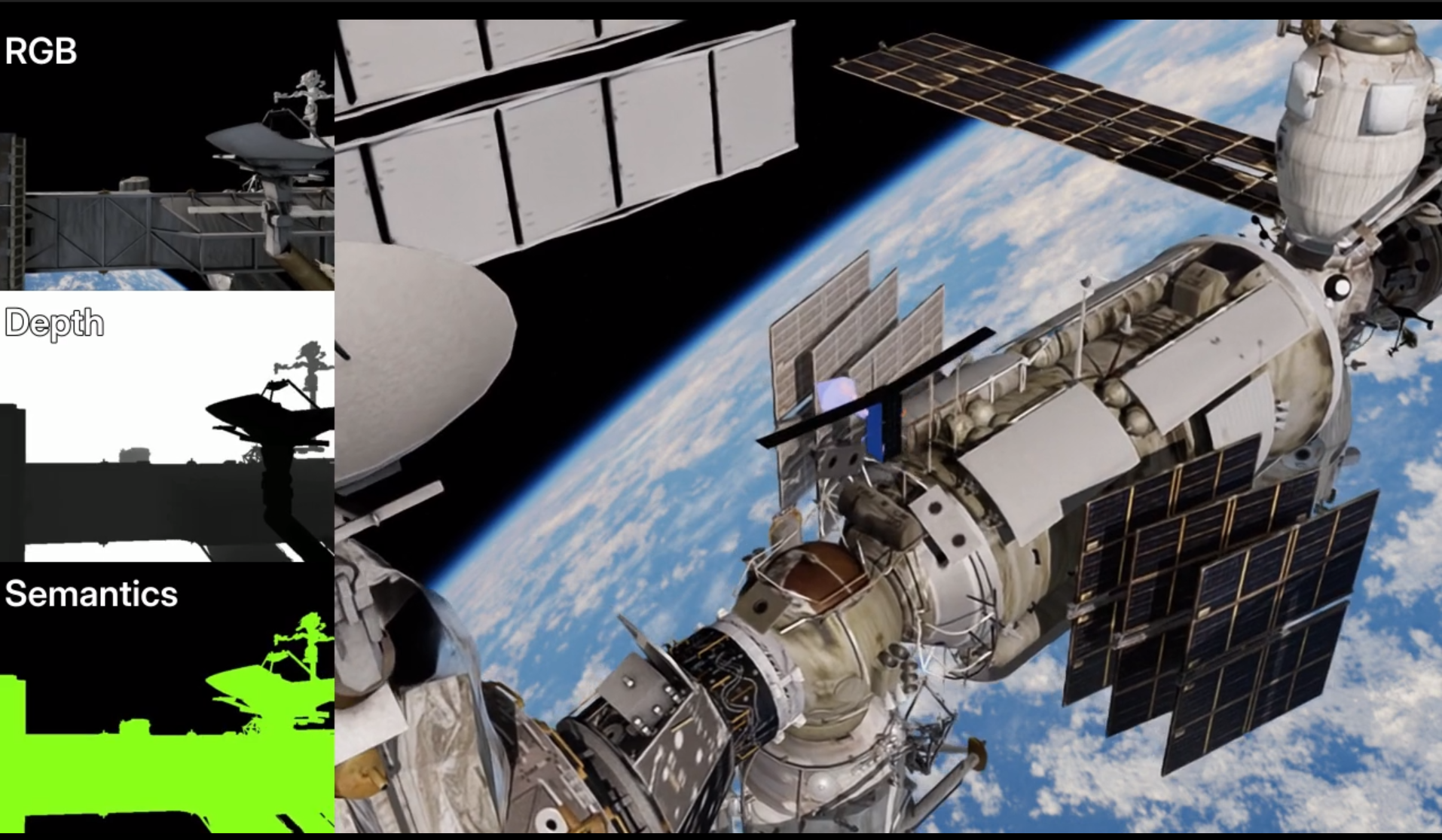}
    }\hfill
    \subfloat{%
         \includegraphics[width=0.9\linewidth,height=0.4\linewidth]{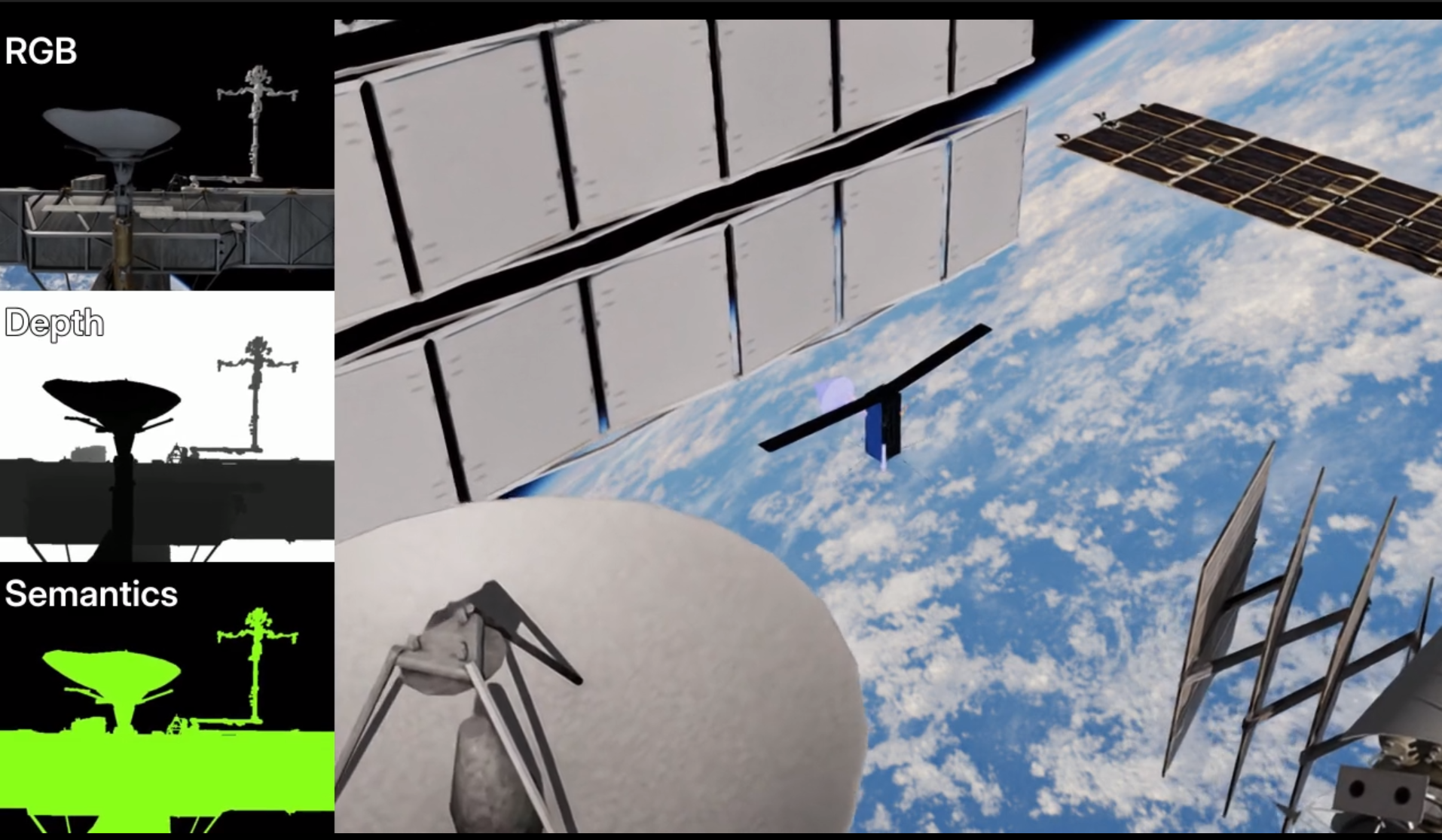}
    }\hfill
    \subfloat{%
         \includegraphics[width=0.9\linewidth,height=0.4\linewidth]{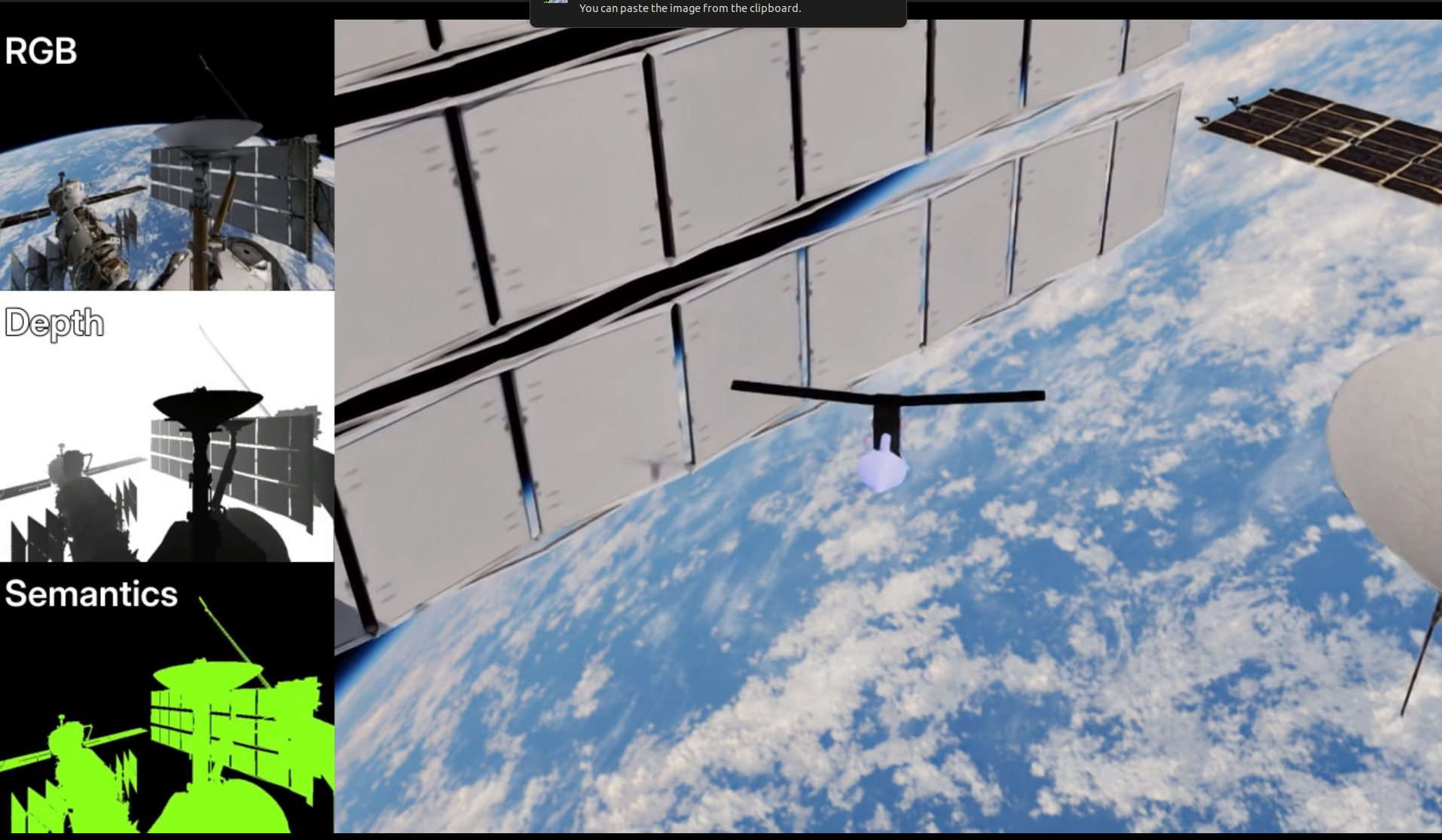}
    }
    \small
    \caption{Sequence of inspection frames showing the spacecraft trajectory around the ISS.}

    \label{fig:traj_sequence_iss}
\end{figure*}

To demonstrate the real-world applicability of the proposed system, we deploy the best DreamerV3 policy (fine-tuned on the capsule trajectory) in three different orbital environments:

\begin{itemize}
    \item Lunar Gateway, Fig.~\ref{fig:traj_sequence_gateway}
    \item Venus Express, Fig.~\ref{fig:traj_sequence_venus}
    \item International Space Station (ISS), Fig.~\ref{fig:traj_sequence_iss}
\end{itemize}

For each target, we present two visualizations: (i) a 3D animation render showing the spacecraft motion around the target, and (ii) the spacecraft camera view during inspection. 
The full set of visual figures is provided in Appendix~\ref{appendix:visuals}.

\section{Conclusion} 
In this work, we introduce \textbf{RL-AVIST}, a reinforcement learning framework for autonomous visual inspection of space assets. We show that model-based RL (DreamerV3) outperforms model-free baselines (PPO, TD3) in both sample efficiency and final performance, establishing it as the most effective approach for long-horizon orbital inspection tasks. Using DreamerV3 as the reference model, we demonstrated its ability to track a wide range of inspection trajectories, including capsule, circle, rectangle, lemniscate, Lissajous, and spiral patterns, highlighting robustness to structured and complex geometric maneuvers. These results confirm the potential of learning-based control to deliver both adaptability and precision in on-orbit operations. Future work will focus on closing the sim-to-real gap by integrating RL-AVIST with hardware-in-the-loop experiments and investigating perception-based policies that fuse visual and dynamic cues for real-time spacecraft inspection.

\small
\newpage
\bibliography{references}

\section*{Appendix}
\subsection{Additional Visual Figures}
\label{appendix:visuals}

This section includes the complete set of visual inspection figures (3D trajectory renders and camera views) for all target environments: Lunar Gateway, Venus Express, and ISS.

\end{document}